\documentclass[10pt,twocolumn,letterpaper]{article}

\usepackage{cvpr}              

\usepackage{algorithm}
\usepackage{algpseudocode}
\usepackage{multirow}

\usepackage[dvipsnames]{xcolor}

\definecolor{cvprblue}{rgb}{0.21,0.49,0.74}
\usepackage[pagebackref,breaklinks,colorlinks,citecolor=cvprblue]{hyperref}

\def\paperID{15589} 
\def\confName{CVPR}
\def\confYear{2024}

\title{mTREE: Multi-Level Text-Guided Representation End-to-End Learning for Whole Slide Image Analysis}

\author{Quan Liu, Ruining Deng, Can Cui, Tianyuan Yao, Vishwesh Nath, Yucheng Tang, Yuankai Huo \\
Vanderbilt University \\
{\tt\small yuankai.huo@vanderbilt.edu}
}

\begin{document}
\maketitle
\begin{abstract}

Multi-modal learning adeptly integrates visual and textual data, but its application to histopathology image and text analysis remains challenging, particularly with large, high-resolution images like gigapixel Whole Slide Images (WSIs). Current methods typically rely on manual region labeling or multi-stage learning to assemble local representations (e.g., patch-level) into global features (e.g., slide-level). However, there is no effective way to integrate multi-scale image representations with text data in a seamless end-to-end process. In this study, we introduce Multi-Level Text-Guided Representation End-to-End Learning (mTREE). This novel text-guided approach effectively captures multi-scale WSI representations by utilizing information from accompanying textual pathology information. mTREE innovatively combines – the localization of key areas (\textbf{``global-to-local"}) and the development of a WSI-level image-text representation (\textbf{``local-to-global"}) – into a unified, end-to-end learning framework. In this model, textual information serves a dual purpose: firstly, functioning as an attention map to accurately identify key areas, and secondly, acting as a conduit for integrating textual features into the comprehensive representation of the image. Our study demonstrates the effectiveness of mTREE through quantitative analyses in two image-related tasks: classification and survival prediction, showcasing its remarkable superiority over baselines. 

\end{abstract}    
\section{Introduction}
\label{sec:intro}


Analyzing Whole Slide Images (WSIs) is a critical aspect of medical imaging research. WSIs are digitized scans of histological samples captured at multiple magnifications, preserving both the overarching view and intricate microscopic details. From a broader perspective, WSIs offer a macroscopic overview of tumor distribution throughout the entire digital slide. This allows for the study of spatial relationships and general tumor traits. At the same time, WSIs enable detailed inspections of cell and tissue structures at the microscopic scale, significantly enhancing diagnostic accuracy and advancing the field of histopathology research~\cite{pantanowitz2011review}.



Over recent years, computer vision has become increasingly crucial and successful in analyzing Whole Slide Images (WSIs), tackling the challenges of handling their super high-resolution ($>10^9$ gigapixels) for tasks like image classification, object detection, and segmentation. Unlike these tasks, our study confronts the unique challenge of performing multi-modal representation learning (involving both image and textual data) for WSIs, focusing on modeling information across multiple scales in both images and text. Textual data in this context can describe both broad and detailed aspects of multi-scale WSIs. A key unresolved question is how to effectively learn representations that encompass both global and local features.

Current methods often rely on manually labeling areas of interest for local representation or using multi-stage learning to merge these local features into a global representation. Yet, these approaches typically fall short of seamlessly integrating multi-scale image representations with text data in an end-to-end process. 

In this paper, we propose Multi-Level Text-Guided Representation End-to-End Learning (mTREE), an innovative text-guided method that effectively captures multi-scale image representations through the use of accompanying textual pathology data. mTREE uniquely blends two formerly separate processes – the identification of crucial areas (\textbf{``global-to-local"}) and the creation of a WSI-level image-text representation (\textbf{``local-to-global"}) – into a unified, end-to-end learning framework.



\begin{figure*}
  \centering
   \includegraphics[width=\linewidth]{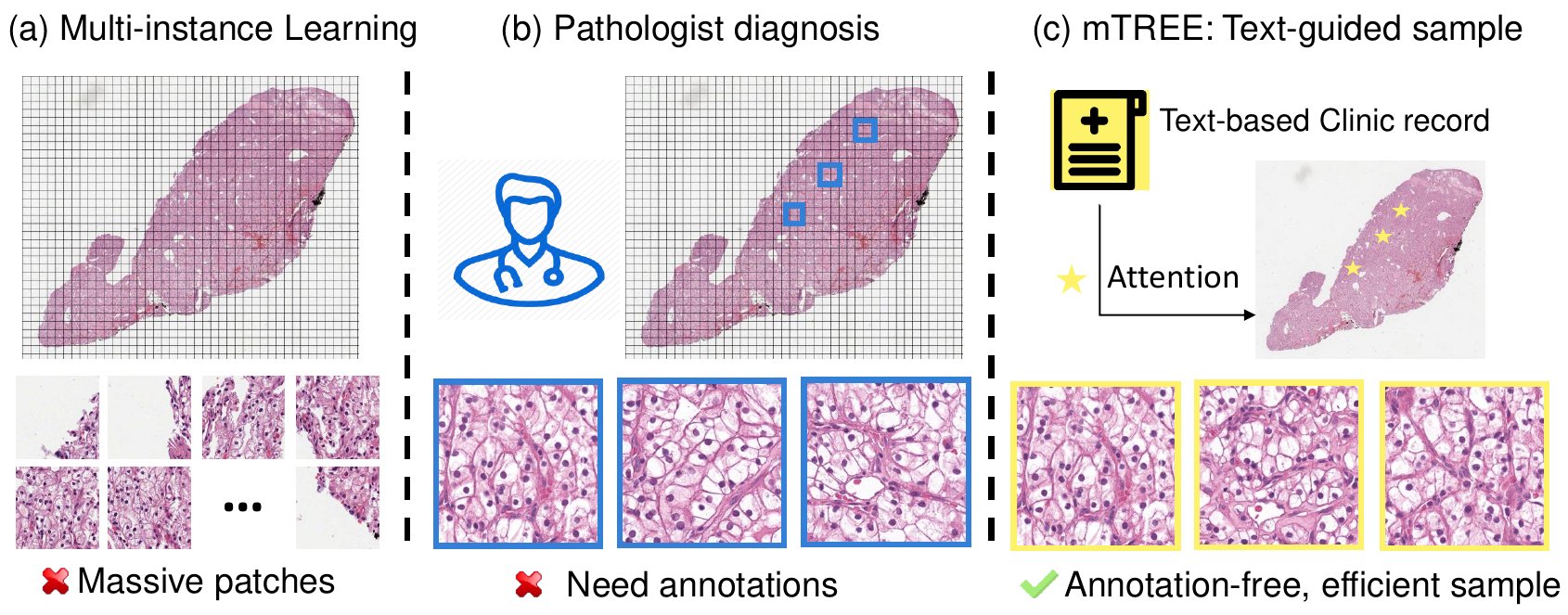}

   \caption{\textbf{Comparison between multi-instance learning, pathologist diagnosis, and our proposed mTREE.} (a) Traditional multi-instance learning needs to process all patches without patch selection. (b) Pathologists in the diagnosis process focus on the most essential patches selected by manual efforts. (c) Our proposed mTREE generates text-guided attention to sample efficiently without manual annotation. }
   \label{fig:idea}
\end{figure*}


While text-based clinical records are consistently available, they haven't been fully utilized in multi-modal representation learning for high-resolution images. Our goal is to develop an algorithm that leverages these textual records to guide the selection of diagnostic patches and aggregate WSI representations without the need for manual annotations. We hypothesize that there is an inherent correlation between the text and image domains; for a given WSI, the clinical text can provide criteria for selecting WSI patches and extracting features. 

Mathematically, Given a WSI $X$, considered as a set of image patches $\left \{x_{i} \right \}_{i=1}^{N}$ where $x_{i} \in X$, and a corresponding label $Y$ for WSI $X$, our approach involves training two mappings. The first is for patch selection $A : \left \{x_{i} \right \}_{i=1}^{N} \rightarrow \left \{x_{i} \right \}_{i=1}^{K}, (K \ll N)$, and the second is for feature abstraction $E : \left \{x_{i} \right \}_{i=1}^{K} \rightarrow Y$. Textual data provides guidance for both mappings: optimizing the selection from the original WSI set and consolidating patch-level features into a comprehensive WSI feature. This dual mapping requires a multi-level approach for text-guided analysis, executed in an end-to-end fashion. In the first mapping, patch selection from $\left \{x_{i} \right \}_{i=1}^{N}$ depends on each patch's relevance to the final prediction, determined by an attention map that scores each patch's importance. Due to the extensive size of WSIs, the attention map is initially learned on lower-resolution images and then mapped to high-resolution images based on coordinate relationships. This approach allows the model to process only a fraction of patches ($K$ out of $N$) when $K \ll N$. In the second mapping, the model uses the features extracted from the selected patches, with the text feature identifying and amalgamating the most pertinent features (those with smaller feature distances) into a unified WSI representation.

In summary, our study introduces a text-guided representation learning method aimed at improving efficiency and extracting features from vital image regions, thereby eliminating the confusion caused by unnecessary image patches. Our approach does not require image annotations from pathologists. Instead, it leverages text descriptions from clinical records to guide the learning process of WSI representations at multiple levels in an integrated, end-to-end manner. We have applied our method to various applications, including image classification and survival prediction across multiple WSI datasets, and have compared it with previous approaches based on Multiple Instance Learning (MIL) models.

The key contributions of our work are fourfold:
\begin{itemize}
    \item We present the first efficient visual-language model for gigapixel WSIs, operating in a seamless end-to-end fashion.
    \item We utilize text information to optimize learning strategies across multiple levels.
    \item Our pipeline is weakly supervised at the WSI level, eliminating the need for patch-level annotations from pathologists.
    \item Our model offers explainability by providing visualizations at different levels, such as attention maps and significant patches.
\end{itemize}


\section{Related Work}
\label{sec: related work}

\subsection{Multi-instance learning}

For pathology image analysis, Multi-Instance Learning (MIL) has emerged as a prominent paradigm, offering a robust framework to address challenges associated with the lack of patch-wise annotation of pathological images~\cite{yao2020whole, hou2016patch, lu2021ai}. Different from the supervised learning method on patches, \cite{edwards2016towards, zaheer2017deep} regard the pathology image as a collection of multiple instances or regions, each potentially harboring critical information for diagnostic or prognostic purposes. This approach allows the model to operate on bags of instances for weakly-supervised learning~\cite{carbonneau2018multiple}. MIL has demonstrated its efficacy in capturing nuanced spatial relationships~\cite{zhao2020predicting} and patterns~\cite{wu2022deepgcnmil} within pathology images, accommodating the inherently diverse nature of tissue structures~\cite{shao2023lnpl, campanella2019clinical} and cellular compositions~\cite{kraus2016classifying}. Upon the set-based concept, Ilse et al.~\cite{ilse2018attention} apply the attention mechanism to Whole Slide Images (WSIs). In a similar vein, Yao et al.~\cite{yao2020whole} integrated attention-based MIL into clustered phenotypes, yielding promising outcomes. Furthermore, \cite{campanella2018terabyte} validated the MIL performs well on large-scale WSI datasets.

\subsection{Attention sampling}
Performing analysis on large images, attention sampling~\cite{xu2015show} has emerged as a powerful technique to efficiently process extensive visual datasets by selectively focusing computational resources on regions of interest. Attention sampling aims to address the challenges posed by the sheer scale of high-resolution images, where the majority of the content may be irrelevant to the specific task at hand~\cite{zheng2019looking}. Attention sampling leverages mechanisms inspired by human visual attention~\cite{hassanin2022visual}, directing computational resources toward salient regions while bypassing less informative areas. Notable approaches include the integration of attention mechanisms within convolutional neural networks (CNNs)~\cite{wang2023attention,xue2022progress, wang2021wide} to dynamically weigh the importance of different image regions. Additionally, attention sampling strategies, such as region-based methods and attention-guided sampling, have been proposed to enhance computational efficiency in tasks such as object detection~\cite{cheng2021weighted, li2019motion}, image classification~\cite{dong2022weighted, wang2017residual}, and segmentation~\cite{kulharia2020box2seg, shi2022dense}.

\subsection{Visual language model in WSI analysis}

The integration of visual language models (VLM) has emerged as a cutting-edge approach, revolutionizing the interpretation of large-scale pathological images. Visual language models combine the strengths of natural language processing (NLP) and computer vision, enabling a comprehensive understanding of complex visual information assisted by knowledge from multiple domains. Unlike fine-tuning, VLM is based on prompt prediction in the template, as seen in CLIP~\cite{radford2021learning} and CoOp~\cite{zhou2022learning}. The trained language model has a strong capability in knowledge and zero-shot learning~\cite{brown2020language}. By leveraging pre-trained language models such as BERT~\cite{devlin2018bert} and adapting them to the unique challenges of WSI, researchers have achieved remarkable strides in capturing contextual relationships and hierarchical structures within pathology images~\cite{huang2023visual}. These models empower the extraction of meaningful features and semantic understanding, enhancing the interpretability of WSIs for tasks such as image classification, tumor detection, and prognosis prediction~\cite{lu2023towards}. 

\section{Methods}
\label{sec: methods}

\begin{figure*}[h]
  \centering
   \includegraphics[width=\linewidth]{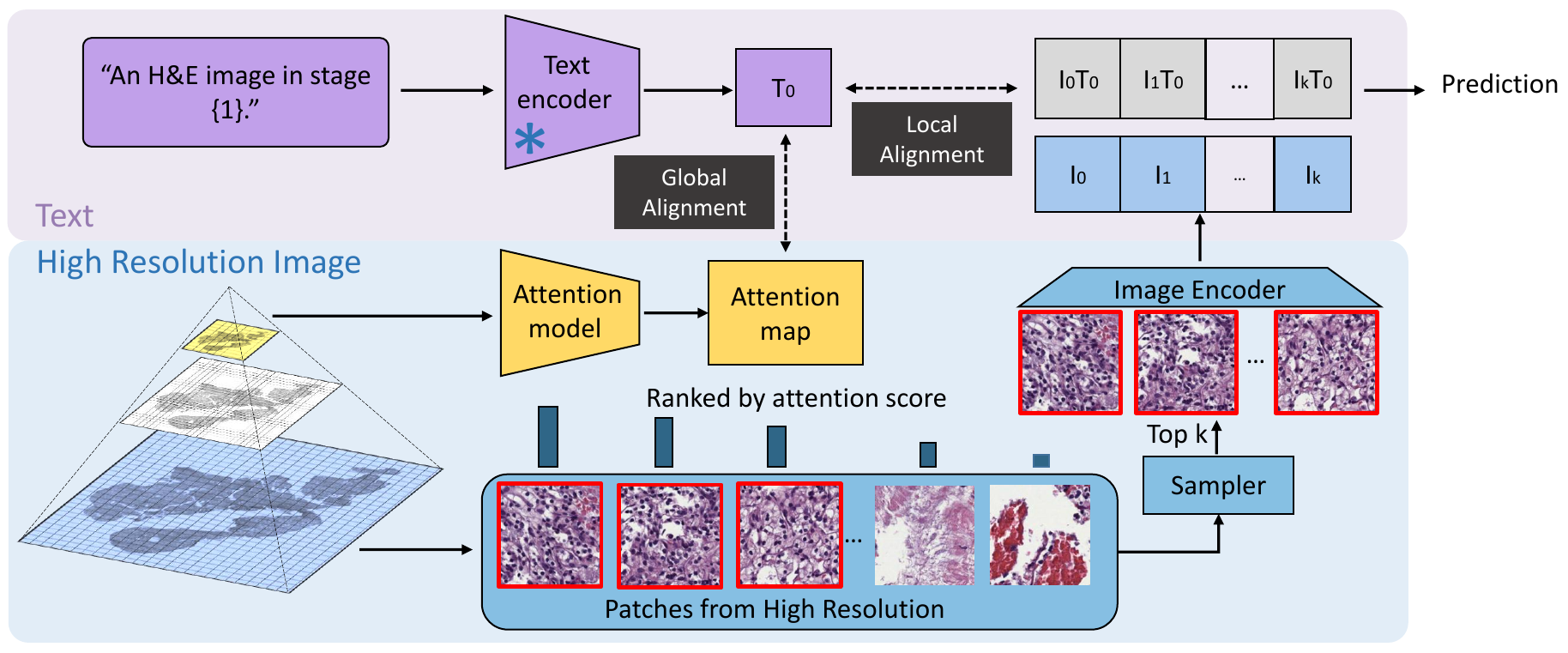}

   \caption{\textbf{This figure demonstrates the proposed mTREE pipeline.} The upper panel shows the text process flow. The text encoder is frozen with pre-trained weights. Text feature $T_{0}$ is used for global alignment and alignment of image patch features. The lower panel shows the WSI analytic flow. The attention model learns an attention map from the WSI in low resolution. The attention map aligns with the text feature $T_0$. The image patches tiled up from high-resolution WSI are ranked by attention score. The image features $I_0, I_1... I_{k}$ abstracted from image patches with higher attention scores are aggregated with text feature $T_0$.}
   \label{fig:pipeline}
\end{figure*}

\begin{figure*}[h]
  \centering
   \includegraphics[width=\linewidth]{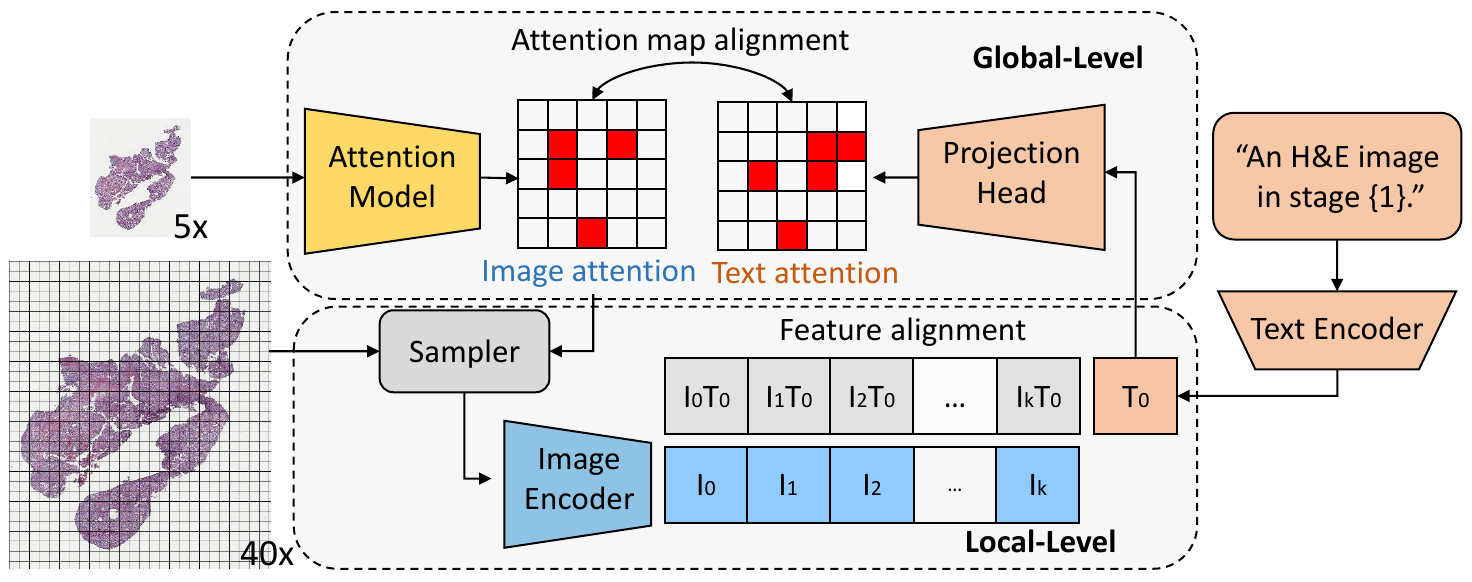}

   \caption{\textbf{This figure presents the principle of multi-level text guidance.} Global-level text guidance (upper panel) aligns the attention map from images and text. Image attention map is learned from low-resolution WSI, while text attention is projected from the text feature $T_0$. Local-level text guidance (lower panel) performs patch selection by computing the cosine similarity distance to the text feature $T_0$ and aggregates features from both image and text.}
   \label{fig:multilevel}
\end{figure*}

As illustrated in \cref{fig:pipeline}, our goal is to learn the multi-level mapping from gigapixel WSI $X$ to the representation $R$ guided by text prompt $T$. To better describe the model learning on gigapixel images, we regard high-resolution images (base layer in image pyramid in \cref{fig:pipeline}) as a collection of image patches $\left \{x_{i} \right \}_{i=1}^{N}$ $(x_{i} \in X)$. To efficiently learn the WSI representation, the global-level alignment learns an attention map $M$ from WSI in low resolution $W_{low}$. The global alignment on $M$ is between image attention map $M_w$ and text attention map $M_t$. The image patch collection $\left \{x_{i} \right \}_{i=1}^{N}$ is ranked by the attention score in attention map $M$. Top $K$ image patches $\left \{x_{i} \right \}_{i=1}^{K}$ with highest attention score are selected by the sampler. Image patch features $\left \{I_{i} \right \}_{i=1}^{K}$ is abstracted by image encoder $E$ from top $K$ image patches $E(x_i)$. The image feature $I_i$ is selected by the cosine similarity with $T_0$. The WSI representation $R$ is aggregated by the $I_i$ weighted by distance $I_0 \times T_0$. At the core of our approach is the idea of multi-level alignment from supervision contained in natural language and the end-to-end training manner to coordinate the region localization and feature extraction.

\subsection{Global alignment}
The attention map can, in theory, learn the significance of the region playing a role in the final prediction. However, for the high-resolution WSI normally in size of $100,000\times100,000$, a gigapixel attention map is too large for attention model capacity to learning. Considering the histopathology image composed of regions of different tissue types, the region distribution of WSI makes attention map learning even harder to match the target prediction. Since learning the attention map for the high-resolution image is not valid, we can learn the attention map in low-resolution WSI to reduce the training burden and localize the rough essential region. To further optimize the attention map, the global-level alignment is between text feature $T_0$ and low-resolution WSI $X_{low}$. The attention mappings on the text side $A_t: T_0 \rightarrow M_t$ 
and WSI side $A_w: X_{low} \rightarrow M_w$ provides the WSI attention map $M_w$ and the text attention map $M_t$. The mapping $A_t$ is achieved by a projection head, composed of multiple fully connected layers. We use MSELoss as attention loss $\mathcal{L}_{Attn}$ to match the $M_w$ with $M_t$ as expressed in \cref{eq:loss_attn}. 
\begin{equation}
  \mathcal{L}_{Attn} = \frac{1}{N} \sum_{i=1}^N (M_w^i - M_t^i)^2
  \label{eq:loss_attn}
\end{equation}
To prevent the attention from collapsing and to encourage non-zero values in the attention map, the sparsity-inducing regulation term $\mathcal{L}_{Sparse}$ is applied to both $M_t$ and $M_w$, penalizing overly sparse attention maps following \cref{eq:loss_sparse}.
\begin{equation}
  \mathcal{L}_{Sparse} = \frac{1}{N} \sum_{i=1}^{N} |M_t| + \frac{1}{N} \sum_{i=1}^{N} |M_w|
  \label{eq:loss_sparse}
\end{equation}
To keep the low-resolution WSI in the same size from the same magnification, the original image is center-cropped before being fed into the attention model. To prevent the attention from focusing only on the cropped image boundary, the $\mathcal{L}_{Boundary}$ is defined by gradients of the attention map with respect to the x-axis ($G_x$) and y-axis ($G_y$). 
The $\mathcal{L}_{Boundary}$ is expressed in \cref{eq:loss_boundary} where $G_x(M_w)_i$ is the $i$-th element in the gradient of $M_w$ with respect to x-axis and $G_y(M_w)_i$ is the $i$-th element in the gradient of $M_w$ with respect to y-axis.
\begin{equation}
  \mathcal{L}_{Boundary} = \lambda \left( \sum_{i=1}^{N} |G_x(M_w)_i| + \sum_{i=1}^{N} |G_y(M_w)_i| \right)
  \label{eq:loss_boundary}
\end{equation}
The full loss function for attention map global alignment is:
\begin{equation}
  \mathcal{L}_{Global} = \mathcal{L}_{Attn} + \mathcal{L}_{Sparse} + \mathcal{L}_{Boundary}
  \label{eq:loss_attn_full}
\end{equation}

\subsection{Local alignment}
The attention map $M$ provides the criterion score for image patch significance. The top $K$ patch selected from $\left \{x_{i} \right \}_{i=1}^{N}$ provides the most essential patches for the prediction. As shown in \cref{fig:multilevel} lower panel, essential patch collection $\left \{x_{i} \right \}_{i=1}^{K}$ selected by the sampler is fed into the image encoder to generate the image feature $\left \{I_{i} \right \}_{i=1}^{K}$. Because the image patches and text are both encoded into feature space, the similarity of text feature $T_0$ with each image feature candidate $I_i$ can be computed and ranked based on \cref{eq:similarity}
\begin{equation}
  \text{Cosine Similarity}(\left \{I_{i} \right \}_{i=1}^{K}, T_0) = \left \{\frac{I_{i} \cdot T_0}{\|I_{i}\| \|T_0\|} \right \}_{i=1}^{K}
  \label{eq:similarity}
\end{equation}
Image feature $\left \{I_{i} \right \}_{i=1}^{J}$ are selected based on the cosine similarity with the text feature $T_0$. The final representation for WSI is aggregated with the selected image features, as shown in \cref{eq:final_rep}.

\begin{equation}
  R=\sum_{i=1}^{J}(\frac{I_{i} \cdot T_0}{\|I_{i}\| \|T_0\|}\cdot I_i)
  \label{eq:final_rep}
\end{equation}

\subsection{End-to-end training with image sampler}
The multi-level alignment under text guidance is built in an end-to-end manner to provide the advantages from two aspects: (1) Coordinate the optimization on the attention map and feature selection and aggregation. (2) Provide the backpropagation path from the final prediction to the attention map for the sampler. Representation learning with attention maps often involves a two-stage training process, where the first stage focuses on learning a representation, and the second stage incorporates attention mechanisms as in \cite{xu2014deep, vaswani2017attention, mnih2014learning, ng2015beyond}. Insufficient end-to-end optimization potentially leads to suboptimal integration of attention and representation learning. To better describe the end-to-end manner, the pseudocode of mTREE is provided.

\begin{algorithm}
\caption{Pseudocode for mTREE implementation}
\label{alg:two_stage_learning}

\begin{algorithmic}[1]
\State \textbf{Input:} Gigapixel WSI $X$, Text Prompt $T$
\State \textbf{Output:} WSI Representation $R$

\State Learn attention map $M_w$ from low-resolution $X_{low}$
\State Align image attention map $M_w$ and text attention map $M_t$ with attention loss
\State Rank image patches $\left \{x_{i} \right \}_{i=1}^{N}$ using attention score in $M_w$
\State Select top $K$ image patches $\left \{x_{i} \right \}_{i=1}^{K}$ with highest attention score

\State Extract image features $\left \{I_{i} \right \}_{i=1}^{K}$ with image encoder $E$
\State Select image feature $I_i$ based on cosine similarity with $T_0$
\State Aggregate WSI representation $R$ using $I_i$ weighted by distance $I_0 \times T_0$

\end{algorithmic}
\end{algorithm}

\section{Experiments}

\begin{table*}[h]
  \centering
  \begin{tabular}{lccccc}
    \toprule
    Method        & Inputs   &  Patch \#  &  Image encoder  &   Acc (grade)  &   C-Index (survival) \\
    \midrule
    PathomicFusion~\cite{chen2020pathomic}  & diagnostic regions (\textcolor{red}{\textbf{manual}})   &  ~20      &  ResNet-50       &  N/A   &   63.1 \\
    
    AttenDeepMIL~\cite{ilse2018attention}  & diagnostic regions (\textcolor{red}{\textbf{manual}})     &  ~20        &  ResNet-50       &   60.9  &   61.5\\
    \midrule
    AttenSample~\cite{katharopoulos2019processing}   & raw WSI (\textcolor{blue}{\textbf{automatic}}) &  1          &  ConvNet       &   49.1 &   55.4\\
    AttenDeepMIL~\cite{ilse2018attention}  & raw WSI (\textcolor{blue}{\textbf{automatic}}) &  100        &  ResNet-50       &   51.0 &   58.8\\
    CLAM~\cite{lu2021data}          & raw WSI (\textcolor{blue}{\textbf{automatic}}) &  $>$5000    &  ResNet-50       &   57.5 &   60.1\\
    mTREE (Ours)        & raw WSI (\textcolor{blue}{\textbf{automatic}}) &  10        &  ResNet-50       &   63.1 &   63.2\\
    mTREE (Ours)        & raw WSI (\textcolor{blue}{\textbf{automatic}}) &  20        &  ResNet-50       &   \textbf{64.7} &   \textbf{65.1}\\
    \bottomrule
  \end{tabular}
  \caption{Cancer grade classification and survival prediction results on KIRC dataset. Acc represents the accuracy of grade classification, while the C-Index evaluates the survival prediction performance.}
  \label{tab:KIRC}
\end{table*}

\subsection{Data description}
To substantiate our proposed text-guided representation learning approach, integrating histological and text features, we sourced glioma and clear cell renal cell carcinoma data from the TCGA, a comprehensive cancer data consortium housing paired high-throughput text in clinic records and diagnostic whole slide images. This dataset is enriched with ground-truth survival outcomes and histologic grade labels. For both the TCGA-KIRC (519 WSIs) and TCGA-GBMLGG (1589 WSIs) projects, region-of-interests (ROIs) from diagnostic slides are provided by \cite{chen2020pathomic}. For clear cell renal cell carcinoma in the TCGA-KIRC project, 512$\times$512 ROIs from diagnostic whole slide images are provided as the diagnostic region. This yielded 3 ROIs per patient (512$\times$512 at 40$\times$ magnification) for 417 patients, resulting in a total of 1251 images. For the TCGA-GBMLGG project, 1024$\times$1024 region-of-interests (ROIs) from diagnostic slides are leveraged. The WSI data is publicly available on the TCGA database~\cite{tomczak2015review}.

\subsection{Data preprocessing}
Both WSI image and text data require preprocessing before feature extraction. The preprocessing for both datasets follows the same strategy. 

\textbf{WSI image data.}  The input image data for our pipeline is provided from two levels: low-resolution images and high-resolution images. The low-resolution images are from 5x magnifications in the WSI pyramid structure. To ensure low-resolution images in the same size and scale, we center-crop the $5,000 \times 5,000$ patches from the low-resolution images. All $5,000 \times 5,000$ patches are then resized to $500 \times 500$. 

\textbf{Text data.}  Follow the design in \cite{huang2023visual, lu2023towards}, text information is composed of templates and prompts curated from the clinical records. In our experiments, the paragraph related to "survival time" and the "cancer grade" are used as text information.

\subsection{Network architectures}
We adopt the representation learning flow from the "low-resolution" to "high-resolution", as proposed in Attention-Sampling~\cite{katharopoulos2019processing}, which has shown impressive results on megapixel image analysis. Based on the "low-resolution" to "high-resolution" strategy, our mTREE pipeline is composed of three parts: image feature analysis models, text feature analysis models, and alignment blocks between image and text.

\textbf{Image feature analysis model.}  Attention model for low-resolution WSI is a Convolution Network (ConvNet) with four convolution layers. We use $3\times 3$ convolution kernel and the channel number of four convolution layers is [8, 16, 32, 1]. The Sampler ranks image patches and selects the top K patches. No learnable parameters in the sampler. ResNet-50 with ImageNet pre-trained weight is used as the image encoder.

\textbf{Text feature analysis model.}
The text encoder is the pre-trained ViT-B/32 used in CLIP~\cite{radford2021learning}. The text encoder is frozen in the training process. 

\textbf{Alignment block.}  The global alignment block includes a projection head composed of two sequential convolution layers to project text feature to text attention map. The local alignment between text features and image features is based on the cosine similarity matrix in the shape of $1 \times K$.

\begin{table*}[h]
  \centering
  \begin{tabular}{lccccc}
    \toprule
    Method        & Inputs   &  Patch \#  &  Image encoder  &   Acc (grade)  &   C-Index (survival) \\
    
    \midrule
    PathomicFusion~\cite{chen2020pathomic}  & diagnostic regions (\textcolor{red}{\textbf{manual}})     &  ~20        &  ResNet-50    &  N/A  &   \textbf{72.4} \\
    AttenDeepMIL~\cite{ilse2018attention}  & diagnostic regions (\textcolor{red}{\textbf{manual}})     &  ~20        &  ResNet-50       &   78.8 &   71.4\\
    \midrule
    AttenSample~\cite{katharopoulos2019processing}   & raw WSI (\textcolor{blue}{\textbf{automatic}}) &  1          &  ConvNet       &   70.4 &   65.4 \\
    AttenDeepMIL~\cite{ilse2018attention}  & raw WSI (\textcolor{blue}{\textbf{automatic}}) &  100        &  ResNet-50       &   70.6 &   63.6\\
    CLAM~\cite{lu2021data}      & raw WSI (\textcolor{blue}{\textbf{automatic}}) &  $>$5000        &  ResNet-50       &   75.3  &   65.7\\
    mTREE (Ours)         & raw WSI (\textcolor{blue}{\textbf{automatic}}) &  10        &  ResNet-50       &   76.5  &   69.0\\
    mTREE (Ours)        & raw WSI (\textcolor{blue}{\textbf{automatic}}) &  20        &  ResNet-50       &   \textbf{79.6} &   70.1\\
    \bottomrule
  \end{tabular}
  \caption{Cancer grade classification and survival prediction results on GBMLGG dataset. 'Acc' represents the accuracy of grade classification, while the C-Index evaluates the survival prediction performance.}
  \label{tab:GBMLGG}
\end{table*}

\subsection{Training details}
We apply our mTREE to two TCGA datasets (KIRC and GBMLGG) on two downstream tasks: grade classification and survival prediction. 

\textbf{Tasks.}
The KIRC dataset has three grades (Stage I, Stage II, and Stage III) for grade classification. The patient's overall survival time in month is used as the label for the survival prediction task. The GBMLGG dataset also has three grades (2, 3, 4) for grade classification. The "Time to last follow-up or death (Month)" is used as the label for survival prediction in the GBMLGG dataset.

\textbf{Hyper-parameters.}
The three most important parameters are evaluated for our proposed mTREE. The first one is the size of the attention map learned in global alignment. Tuned by the projection head channel number and attention model structure, we evaluated attention map size in $123 \times 123$ and $246 \times 246$. The second parameter is the sample number from the attention map (K in \cref{eq:similarity}). Based on the number of patches from the diagnostic region provided by \cite{chen2020pathomic}, approximately 20 patches for each WSI, we evaluated sample number in set {5, 10 20, 50}. The third parameter is the sample number from the image patch features (J in \cref{eq:final_rep}). According to $K \in \left \{ 5, 10, 20, 50\right \}$, we sampled $J \in \left \{ 2, 5, 10, 20\right \}$.

\textbf{Metrics.}
The metric used for grade classification tasks is accuracy (ACC). The ACC evaluates the WSI representation performance on class prediction tasks with discrete labels.

The metric for the survival prediction task is the C-Index. It quantifies the concordance between predicted and observed survival times, with a higher C-index indicating improved predictive accuracy.

\subsection{Baseline experiments for comparison}
To validate the advantages of the proposed mTREE pipeline, the baseline experiments comparisons are compared from three aspects: (1) MIL-based model: we use AttenDeepMIL~\cite{ilse2018attention} as a general MIL-based model and CLAM~\cite{lu2021data}, designed specifically for WSI analysis. (2) Attention Sampling~\cite{katharopoulos2019processing}, and (3) PathomicFusion~\cite{chen2020pathomic}.

\textbf{AttenDeepMIL.}  The implementation of AttenDeepMIL follows the settings in \cite{ilse2018attention}. ResNet-50 with ImageNet-pretrained weights is used as the image encoder. For both the KIRC dataset and the GBMLGG dataset, two patch selection strategies are evaluated: (1) diagnostic region (DR in \cref{sec:results}), and (2) tiled-up image patches from 40x WSI (Origin in \cref{sec:results}).

\textbf{CLAM.}  In \cite{lu2021data}, CLAM provides the implementation of MIL for classification. For a fair comparison, the image encoder is ResNet-50, similar to other baseline methods. CLAM processes all image patches from WSI, except the background patches, normally more than 5,000 patches for a WSI.

\textbf{AttenSample.}  Attention sampling input set has an image in high resolution ($1,500 \times 1,500$) and a low-resolution image rescaled by a ratio of 0.1 ($150 \times 150$). For the TCGA dataset, the high-resolution image is from a 5x magnification WSI and center-cropped in size of $5,000 \times 5,000$. The low-resolution image is rescaled by a ratio of 0.1 to $500 \times 500$.

\textbf{PathomicFusion.} In \cite{chen2020pathomic}, PathomicFusion is a multi-modal fusion method incorporated with image, genomics, and cell graph data. In our experiments, only image data is used for performance comparison.

\section{Results}
\label{sec:results}

Following the experiment settings, two datasets TCGA-KIRC and TCGA-GBMLGG on grade classification and survival prediction are discussed. To optimize the hyperparameter settings, an ablation study of different parameters is evaluated on the survival prediction task on the GBMLGG dataset. To provide the model with explainability, the visualization of the attention map and selected diagnostic region by mTREE is presented.

\subsection{KIRC}

In \cref{tab:KIRC}, we compare the performance of the proposed mTREE with baselines on the TCGA-KIRC dataset for the grade classification task. It is observed that the performance of MIL-based methods improves with an increasing sample number from the WSI patch collection. When processing all patches from the WSI, CLAM achieves an accuracy of 57.5\%. Remarkably, our proposed mTREE outperforms, achieving a superior accuracy of 64.7\% with just 20 sampled patches from the WSI, surpassing even AttenDeepMIL with a diagnostic region.

Moving on to the survival prediction task in \cref{tab:KIRC}, we observe a performance trend similar to the grade classification task. MIL-based methods exhibit better performance with a diagnostic region compared to original image patches. The best performance from the MIL baseline achieves a C-Index of 0.631 when trained on the diagnostic region. Notably, our proposed mTREE demonstrates superior performance (C-Index 0.651) over the baselines, utilizing both original image patches and diagnostic regions.


\subsection{GBMLGG}

In \cref{tab:GBMLGG}, we present the prediction performance of the proposed mTREE and baselines on the grade classification task. Similar to the performance comparison in the grade classification task, MIL-based methods exhibit better performance with a diagnostic region compared to original image patches. The best performance from the MIL baseline achieves an accuracy of 78.8\% when trained on the diagnostic region. Notably, our proposed mTREE outperforms the baselines, achieving a superior accuracy of 79.6\% with both original image patches and the diagnostic region.

\begin{figure*}
  \centering
   \includegraphics[width=0.9\linewidth]{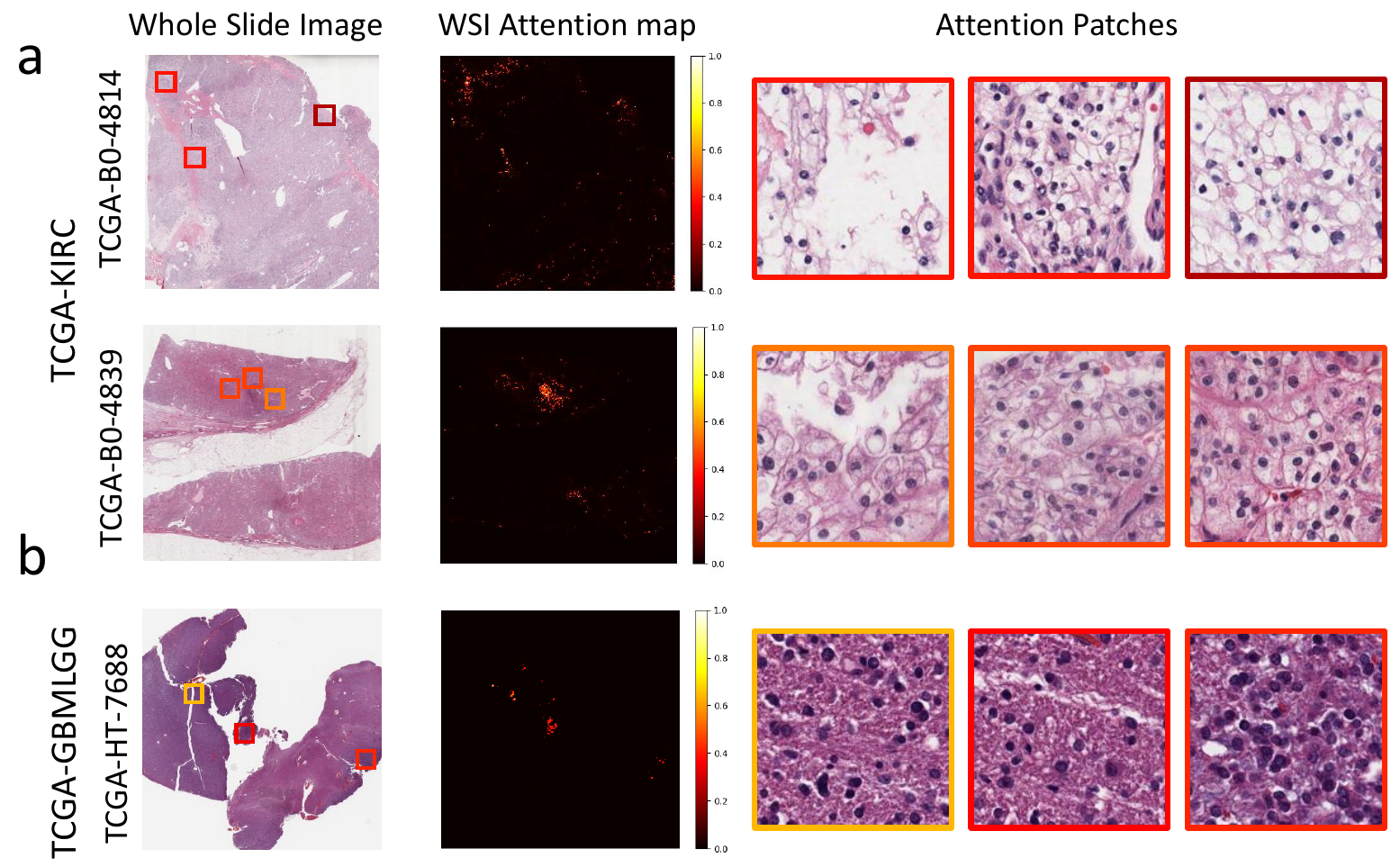}

   \caption{\textbf{This figure presents the visualization of WSI-level attention and the automatically derived diagnosis patches.} For WSIs in the TCGA-KIRC dataset and TCGA-GBMLGG dataset, the attention map (middle panels) is learned from WSI (left panels), highlighting essential tissue regions. Essential image patches (right panels) are selected according to the attention score. The image boundary color indicates the according attention score.}
   \label{fig:visual}
\end{figure*}


In \cref{tab:GBMLGG}, we present the prediction performance of the proposed mTREE and baselines for the survival prediction task. The MIL-based method exhibits better performance with a diagnostic region than with original image patches. The best performance from the MIL baseline achieves a C-Index of 0.724 when trained with the diagnostic region. Notably, our proposed mTREE outperforms the baselines, achieving a better performance (C-Index 0.701) with original image patches.

\subsection{Evaluation for multi-level text alignments}

\begin{table*}[t]
  \centering
  \begin{tabular}{@{}ccccccc@{}}
    \toprule
    Method  &  Global align & Local align   & KIRC ACC  &  KIRC C-Index   &  GBMLGG ACC  &  GBMLGG C-Index   \\
    \midrule
    \multirow{3}{0.9in}{mTREE} &    &    &   49.1 & 55.4 & 70.4 & 65.4 \\
    & \checkmark  &     &  51.0 & 56.3 & 70.7 & 65.2 \\
    & \checkmark  &  \checkmark & \textbf{64.7} & \textbf{65.1} & \textbf{79.6} & \textbf{70.1} \\
    \bottomrule
  \end{tabular}
  \caption{Ablation study for multi-level text alignments is shown in this table. The accuracy of grade classification (ACC) and the survival prediction performance (C-Index) are presented.}
  \label{tab:multilevel}
\end{table*}

In this section, we compare the performance of the proposed mTREE with and without global and local alignment. The results are presented in \cref{tab:multilevel}. From the performance shown in \cref{tab:multilevel}, both global and local alignment contribute to performance improvement in all four tasks. However, global alignment provides a limited contribution (as shown in the second row of \cref{tab:multilevel}). From another perspective, the coordination between local and global alignment underscores the advantages of an end-to-end training approach.

\subsection{Visualization}

The attention map (\cref{fig:visual} middle) obtained through global alignment serves as a crucial tool for improving the interpretability of mTREE in the context of whole-slide image (WSI) analysis. The heightened intensity in the attention map accentuates key regions within the WSI that significantly contribute to the final prediction. In the domain of weakly-supervised learning, these bright regions indicate areas of essential diagnostic relevance. To enhance human understanding, image patches identified as having high diagnostic importance are presented in a zoomed-in view (\cref{fig:visual} right). The color of the image boundary indicates the corresponding attention score. Patches with higher attention scores are deemed more important for the final prediction.

\section{Conclusion}

This paper introduces a novel text-guided representation learning pipeline designed for the efficient processing of Whole-Slide Images (WSIs). Our proposed model, mTREE, seamlessly integrates textual pathology information with WSI features on multiple levels, enabling a comprehensive understanding of the underlying data. Trained in an end-to-end manner, mTREE demonstrates superior performance in both classification and survival prediction across two distinct WSI datasets. Notably, the model exhibits explainability, as evidenced by its capability to visualize attention maps at both the WSI level and specific patches with high diagnostic importance. This fusion of accuracy and interpretability underscores the effectiveness of mTREE in the domain of WSI analysis.

{
    \small
    \bibliographystyle{ieeenat_fullname}
    \bibliography{main}
}


\end{document}